\documentclass[10pt,letterpaper]{article}
\usepackage[top=0.85in,left=1in,right=0.75in,footskip=0.75in]{geometry} 

\usepackage{amsmath,amssymb}

\usepackage{changepage}

\usepackage[utf8x]{inputenc}

\usepackage{textcomp,marvosym}

\usepackage{cite}

\usepackage{nameref,hyperref}

\usepackage[right]{lineno}

\usepackage{microtype}
\DisableLigatures[f]{encoding = *, family = * }

\usepackage[table]{xcolor}

\usepackage{array}

\newcolumntype{+}{!{\vrule width 2pt}}

\newlength\savedwidth

\raggedright
\usepackage[aboveskip=1pt,labelfont=bf,labelsep=period,justification=raggedright,singlelinecheck=off]{caption}

\makeatletter
\renewcommand{\@biblabel}[1]{\quad#1.}
\makeatother

\usepackage{lastpage,fancyhdr,graphicx}
\usepackage{epstopdf}
\usepackage{tikz}
\def\E{\mathbb{E}}
\def\P{\mathbb{P}}
\def\PIT{\mathrm{PIT}}
\newcommand{\minimize}{\mathop{\mathrm{minmize}}}
\newcommand{\st}{\mathop{\mathrm{subject\,\,to}}}

\begin{document}
\vspace*{0.2in}

\begin{flushleft}
{\huge
\textbf\newline{Recalibrating probabilistic forecasts of epidemics} 
}
\newline
\\
Aaron Rumack\textsuperscript{1*},
Ryan J. Tibshirani\textsuperscript{1,2},
Roni Rosenfeld\textsuperscript{1}
\\
\bigskip
\textbf{1} Machine Learning Department, Carnegie Mellon University, Pittsburgh,
PA, USA \\
\textbf{2} Department of Statistics \& Data Science, Carnegie Mellon University,
Pittsburgh, PA, USA \\
\bigskip

*Corresponding author, arumack@andrew.cmu.edu
\end{flushleft}

\section*{Abstract}
Distributional forecasts are important for a wide variety of
applications, including forecasting epidemics. Often, forecasts are
miscalibrated, or unreliable in assigning uncertainty to future events.
We present a recalibration method that can be applied to a black-box
forecaster given retrospective forecasts and observations, as well as
an extension to make this method more effective in recalibrating
epidemic forecasts. This method is guaranteed to improve calibration
and log score performance when trained and measured in-sample. We also prove
that the increase in expected log score of a recalibrated forecaster is
equal to the entropy of the PIT distribution. We apply this
recalibration method to the 27 influenza forecasters in the FluSight
Network and show that recalibration reliably improves forecast accuracy
and calibration. This method is effective, robust, and easy to use as a
post-processing tool to improve epidemic forecasts.

\section*{Author summary}
Epidemics of infectious disease cause millions of deaths worldwide each
year, and reliable epidemic forecasts can allow public health officials
to respond to mitigate the effects of epidemics. However, because
epidemic forecasting is a relatively new field and there is often
insufficient training data, many epidemic forecasts are not calibrated.
Calibration is a desired property of any forecast, and we provide a
post-processing method that recalisrates forecasts. We demonstrate the
effectiveness of this method in improving accuracy and calibration on a
wide variety of influenza forecasters. We also show a quantitative
relationship between calibration and a forecaster's expected score. Our
recalibration method is a tool that any forecaster can use, regardless
of model choice, to improve forecast accuracy and reliability. This
work provides a bridge between forecasting theory, which rarely deals
with applications in domains that are new or have little data, and some recent
applications of epidemic forecasting, where forecast calibration is rarely
analyzed systematically. 


\section{Introduction}
Epidemic forecasting is an important tool to inform the public health
response to outbreaks of infectious diseases. Often, decision makers
can take more effective action with an estimate of the uncertainty in a
forecasted target. For this reason, distributional forecasts are more
desirable than point forecasts. A distributional forecast is a probability
distribution over the target variable and measures the uncertainty in the
prediction, as opposed to a point forecast, which is just a scalar value for
each target and has no measure of uncertainty. A desired property of
distributional forecasts is \textit{calibration}, or reliability between
forecasts and the true distribution of the variable forecasted (a mathematical
definition is given in Section~\ref{sec:methods}). Calibration is one 
of three components of a forecaster's accuracy as measured by any proper 
score \cite{brocker2009}, with better calibration resulting in a better
score. It is therefore important for a forecaster to produce calibrated
forecasts.

Previous work has described general forecasting theory and calibration
and evaluated the calibration of certain forecasts
\cite{dawid1985,gneiting2007,gneiting2007b,hora2004}. Later work has
gone beyond just describing calibration, presenting post-processing
algorithms to recalibrate forecasts that were previously miscalibrated.
Nonparametric techniques for recalibration of ensemble forecasts
include rank histogram correction \cite{hamill1997}, Bayesian model
averaging \cite{raftery2005}, linear pooling \cite{gneiting2013}, and
probability anomaly correction \cite{dool2017}. Brocklehurst et al.\
\cite{brocklehurst1990} provide a nonparametric approach using the
empirical CDF, which can recalibrate any forecast of a scalar target.
Parametric approaches include logistic regression \cite{hamill2004},
extended linear regression \cite{gneiting2005} and beta-transform
linear pooling \cite{gneiting2013}. Wilks and Hamill \cite{wilks2007}
compare the performance of different recalibration techniques for
different meteorological targets with different amounts of training
data. 

Much of the work in recalibration has been applied to weather
forecasting, and thus many of the techniques are not applicable in
other forecasting domains. The most popular weather forecasting models
create a distribution from a series of point predictions, with each
point being the result of a simulation under varying initial
conditions. Many of the existing recalibration methods are defined only
for this type of ensemble forecaster. For example, Bayesian model
averaging assumes that an ensemble forecast is comprised of the same
$N$ forecasts in each observation. This method cannot be extended
trivially to a domain where the forecaster itself outputs a
distribution. Additionally, weather forecasts usually have a plethora
of training data on which to train recalibration methods. For example,
recalibration has been applied to a set of weather forecasts generated
daily from 1979 to at least 2006, almost 10,000 days \cite{hamill2006}.
In settings like these, techniques need not be robust to small amounts
of recalibration training data.

To be clear on nomenclature, throughout this paper, we use the term
\textit{forecast} to refer to the predicted probability distribution of a
variable and the term \textit{forecaster} to refer to an algorithm that produces
a forecast for a variable given a context. Common examples of forecasters are an
algorithm that forecasts the amount of precipitation two days in
advance given current meteorological information, one that forecasts
the price of a certain stock given the stock's historical trend, or one
that forecasts the statewide influenza incidence given historical
incidence data. We also distinguish between \textit{calibration} and 
\textit{recalibration}; calibration refers to the property of a forecaster, and 
recalibration refers to a method whose goal is to make a forecaster
more calibrated. Specifically, recalibration takes as input a set of a
forecaster's forecasts and corresponding observations (``training
data''), and outputs a forecaster which should be more calibrated on a
different set of forecasts and observations (``test data'').

In what follows, we present a generalized approach to
forecast recalibration and show its performance when applied to 
forecasters in the FluSight Network. We demonstrate that
across the diverse set of FluSight forecasters, recalibration consistently
improves not just calibration but accuracy as well.

\section{Methods}
\label{sec:methods}
Consider the following setup. At each $i=1,2,3\ldots$, a forecaster $M$
outputs a density forecast $f_i$ given observations $x_i$ for a continuously
distributed scalar random variable $y_i$ whose true distribution is $h_i$. We 
assume that the corresponding cumulative distribution functions (CDFs)
$F_i$ and $H_i$ are continuous and strictly increasing (for simplicity). The
forecaster $M$ is evaluated according to a proper scoring rule, such as the
Brier score \cite{brier1950} or the logarithmic score \cite{good1952}.   

The goal of a forecaster is to produce ideal forecasts, i.e., to forecast $f_i =
h_i$, the true distribution of $y_i$, for each $i$, though this is usually 
unattainable. We can inspect how close a forecaster is to being ideal with the
distribution of the probability integral transform (PIT) values
\cite{dawid1984}. For each forecast $f_i$ and observed value $y_i$, the PIT is
defined as  
$$
\PIT(f_i, y_i) = F_i(y_i),
$$
where $F_i$ is the CDF of $f_i$. A necessary (but not sufficient)
condition for a forecaster to be ideal is \textit{probabilistic calibration}  
\cite{gneiting2007}:
$$
\frac{1}{N} \sum_{i=1}^N H_i \circ F_i^{-1} (p) \to p \;\; 
\text{as $N \to \infty$}, \;\; \text{for all $p \in (0,1)$}. 
$$
(Here and throughout we interpret convergence in the almost sure sense.) 
An example of a probabilistically calibrated forecaster that is not
ideal is the so-called climatological forecaster, which for each $i$ 
outputs the marginal distribution of $y_i$ over $i=1,2,3,\ldots$. (To make this
concrete, suppose that each $y_i$ is distributed as $\mathcal{N}(\mu_i, 1)$, a
normal distribution with mean $\mu_i$ and variance 1, and each $\mu_i$ itself
follows $\mathcal{N}(0,1)$, then the climatological forecaster simply outputs 
$\mathcal{N}(0,2)$ for each $i$.)

\begin{figure}[p]
    \centering
    \includegraphics[width=0.85\textwidth]{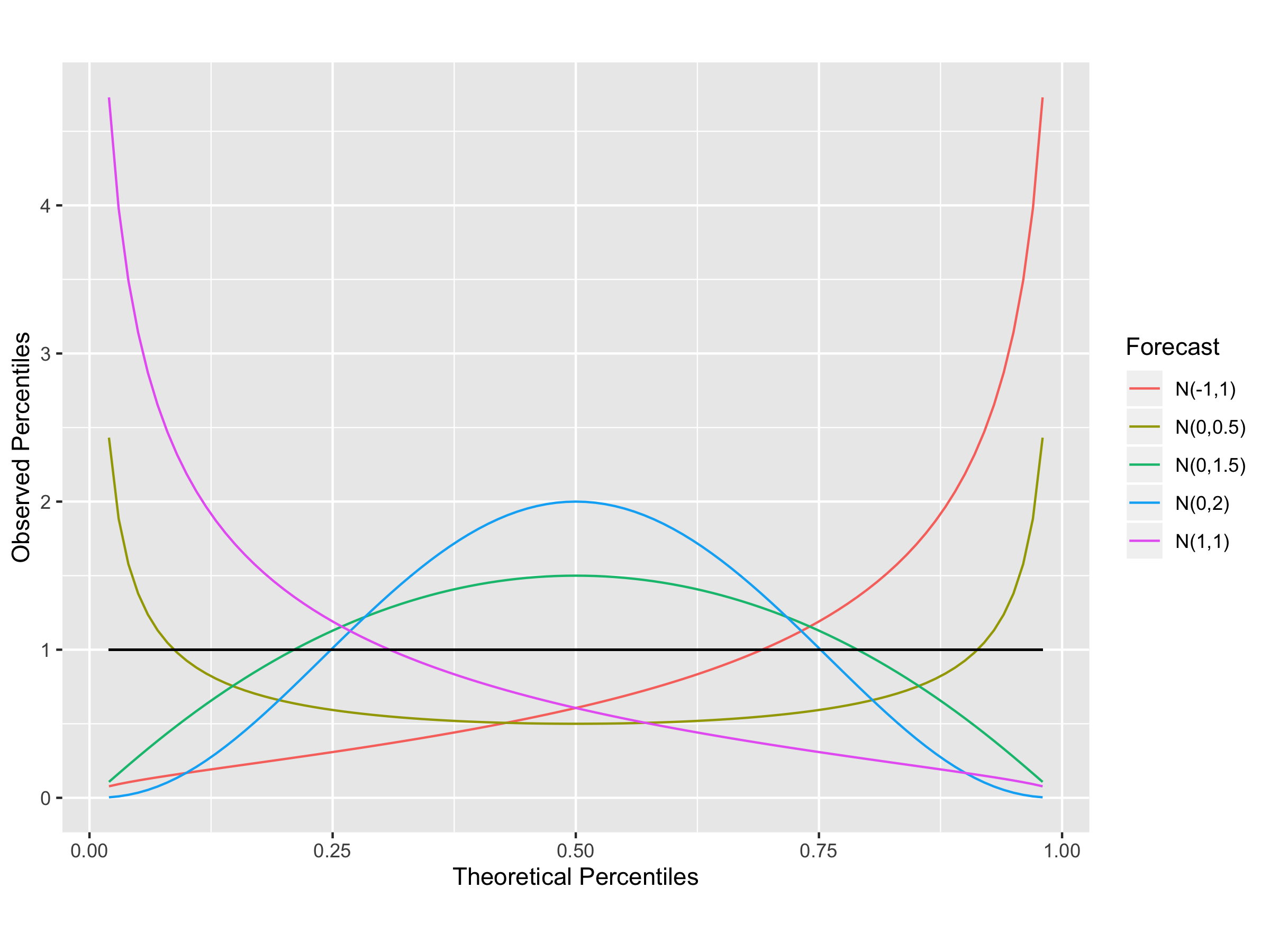}
    \caption{\small Densities of PIT distributions for five sample 
    forecasters, when the true distribution is a standard normal.}
    \label{fig:pit_pdf}

    \bigskip
    \includegraphics[width=0.85\textwidth]{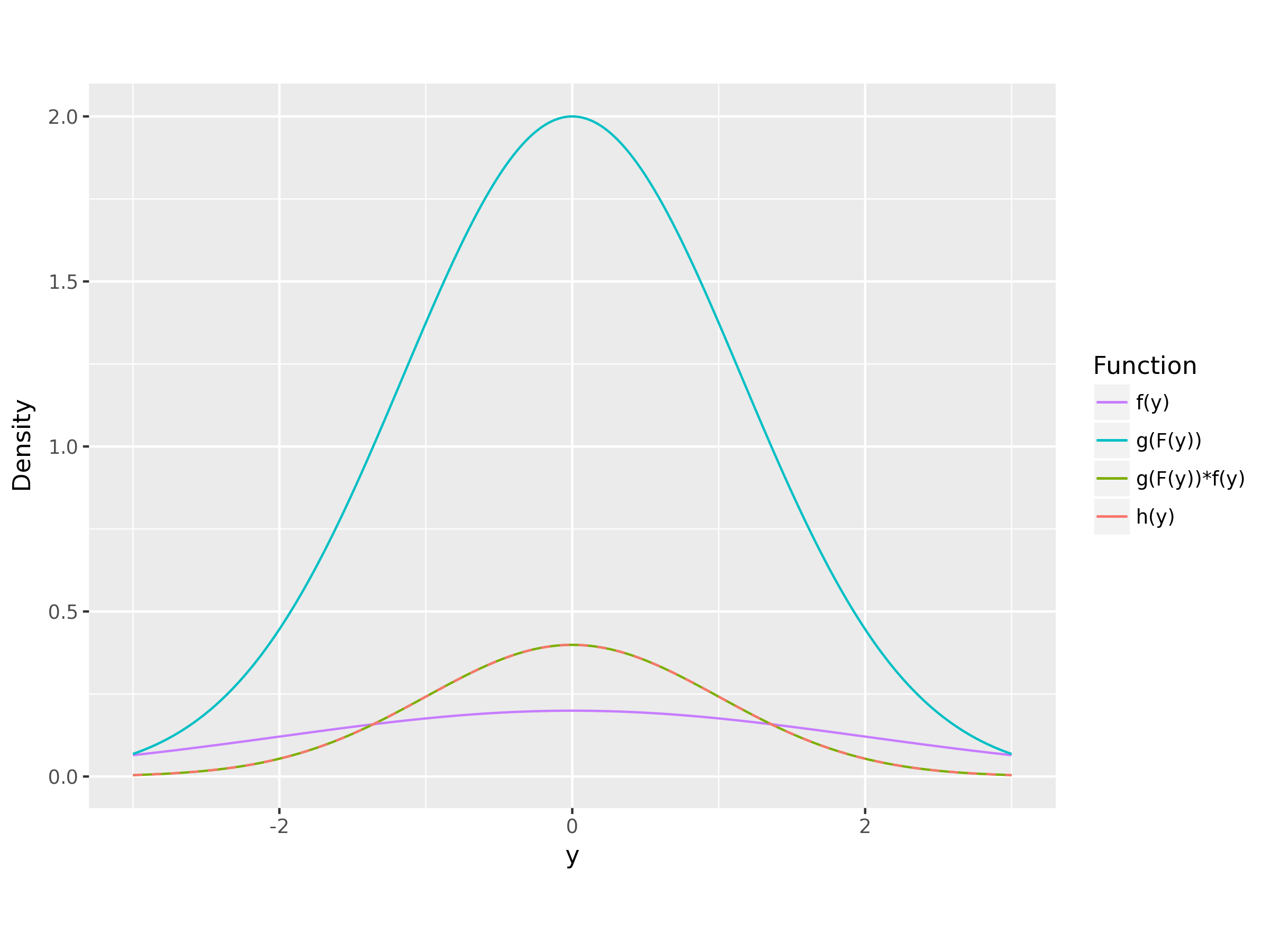}
    \caption{\small An illustration of recalibration. The original,
    underconfident forecast density is $f(y) = \mathcal{N}(0,2)$ while the  
    true density is $h(y) = \mathcal{N}(0,1)$. By calculating the
    PIT density $g$ and producing a recalibrated forecast as the
    product $g(F(y)) \cdot f(y)$, we recover the true $h(y)$.}
    \label{fig:np_ex}
\end{figure}

Note that the PIT distribution of a probabilistically calibrated forecaster is 
close to uniform in large samples. The expected CDF of the PIT distribution
is 
$$
G(p) = \E[\P[F_i(y_i) \leq p]] 
= \E[\P[y_i \leq F^{-1}_i(p)]] 
= \E[H_i \circ F_i^{-1}(p)],
$$
where here $\E$ denotes the sample average operator over $i=1,\ldots,N$. This
expression converges to $p$ as $N \to \infty$ when the forecaster is
probabilistically calibrated. Thus an examination of the distribution of PIT
values---looking for potential deviations from uniformity---serves as a good
diagnostic tool to assess probabilistic calibration. Many use a PIT histogram to
examine the PIT distribution because it is easy to read and understand
\cite{gneiting2007}. For example, if the PIT distribution is bell-shaped, then
the forecaster does not put enough weight in the middle of its distribution and
is underconfident. In general, we can compare the PIT density to the horizontal
line at 1, which corresponds to the uniform density. The greater the deviation
from this line (which can be quantified via Kullback-Leibler divergence from the
uniform distribution to the PIT distribution, or equivalently, negrative netropy
of the PIT distribution), the greater the miscalibration; see
Fig~\ref{fig:pit_pdf} for examples.

Our recalibration method uses $G$ as a CDF-CDF transform. The recalibrated
forecaster, denoted $M^*$, is defined by a recalibrated forecast CDF of
$F_i^*(y) = G(F_i(y))$, for each $i$. By the chain rule, the recalibrated
forecast density is $f_i^*(y) = g(F_i(y)) \cdot f_i(y)$, for each $i$. Thus the 
recalibrated forecast $f_i^*$ is the original forecast $f_i$ weighted by the PIT
density $g$. An illustration of this method is provided in Fig~\ref{fig:np_ex}.
In practice, of course, we do not have access to the true distributions $H_i$,
so we need to estimate $G$. The ultimate estimate of $G$ that we propose in this
paper will be an ensemble (weighted linear combination) of three estimates: a
nonparametric method, a parametric method, and a null method. First, we will
motivate calibration as a tool to increase forecast accuracy, and then, we
explain the individual estimation methods.  

\subsection{Calibration and log score} 
\label{sec:calibration_and_log_score}
In order to quantify how well a forecaster is calibrated, we calculate
the entropy of the distribution of PIT values. As above, $G$ is the CDF
of the PIT distribution of $M$. The entropy of the PIT density $g$ is
defined as
$$
H(g) = -\int_{p=0}^1 g(p)\log g(p) \, dp.
$$
If $M$ is probabilistically calibrated, then (asymptotically, as $N \to \infty$)
the PIT values are uniform and the entropy is zero because $g(p)$ is $1$
everywhere. When the PIT values are not uniform, the entropy is negative. 

Entropy is also useful because it provides an understanding of how
miscalibration penalizes the expected log score, as shown below. First observe
that 
$$
g(p) = \frac{d}{dp}G(p) 
= \frac{d}{dp} \E[ H_i \circ F^{-1}_{i}(p)]
= \E\bigg[ \frac{h_i(F^{-1}_i(p))}{f_i(F^{-1}_i(p))} \bigg],
$$
where the last step assumes the smoothness and integrability conditions on 
$h_i,f_i$ needed to exchange expectation and differentiation (the Leibniz rule). 
Next observe that
\begin{align}
 \nonumber
\E[\log f^*_i(y_i)] - \E[\log f_i(y_i)]
 \nonumber
  &= \E [\log g(F_i(y_i))]  \\
 \nonumber
  &= \E \Bigg[ \int_{-\infty}^\infty \log g(F_i(y_i)) h_i(y_i) \, dy_i \Bigg] \\ 
 \nonumber
  &= \E \Bigg[ \int_0^1 \log g(F_i(F_i^{-1}(p))) \frac{h_i(F_i^{-1}(p))}
    {f_i(F_i^{-1}(p))} \, dp \Bigg] \\   
 \nonumber
  &= \int_0^1 \E \Bigg[ \log g(p) \frac{h_i(F_i^{-1}(p))} {f_i(F_i^{-1}(p))} 
    \Bigg] \, dp \\ 
  \label{eq:calibration_and_log_score}
  &= \int_{p=0}^1 g(p) \log g(p) = - H(g),
\end{align}
where the third line is obtained by a variable substitution, and fourth 
by applying the Leibniz rule again assuming the needed regularity conditions.    

For any forecaster, if the PIT distribution is the same for the training data
and the test data, then the improvement of the recalibrated forecast's log score
can be estimated by estimating the negative entropy of $g$ (note that the
entropy of any distribution on $[0,1]$ is nonpositive). We can explain this
intuitively as well: the more negative $H(g)$ is, the more it indicates that
there is information lying in the structure of $g$ that can be extracted to
improve forecasts.   

\subsection{Nonparametric correction}
Given an observed training set of PIT values for a forecaster, $F_i(y_i)$,
$i=1,\ldots,N$, the empirical PIT CDF is
$$
\hat{G}(x) = \frac{1}{N} \sum_{i=1}^N \mathbb{I} [F_i(y_i)\leq x]. 
$$
As \smash{$\hat{G}$} is discrete, it does not admit a well-defined density, and
hence to use this for recalibration we can first smooth \smash{$\hat{G}$} using
a monotone cubic spline interpolant, and then it will have a bonafide density
\smash{$\hat{g}$}, which is itself smooth (twice continuously differentiable, to 
be precise). Using this for recalibration produces \smash{$f_i^*(y) =
  \hat{g}_i(F_i(y)) \cdot f_i(y)$}.    

In practice, with a large amount of training data, recalibration using the
empirical CDF as described above can be effective. However, with little training 
data, or a lot of diversity within the training data among the distributions of
$y_i$, it can be ineffective for assuring calibration on the test set. This is
in line with the practical difficulties of using nonparametric,
distribution-free methods in general. 

\subsection{Parametric correction}
Gneiting and Ranjan \cite{gneiting2013} present a recalibration method
originally motivated by redistributing weights on the components of an ensemble
forecast, but their method can applied generally to recalibrate any black box
forecaster. Given an observed training set of PIT values, $F_i(y_i)$,
$i=1,\ldots,N$, we fit a beta density \smash{$\hat{g}$} via maximum likelihood  
estimation. This in fact corresponds to the beta transform that maximizes the
log score of the recalibrated forecaster on the training data
\cite{gneiting2013}.  

This parametric model is more resilient to minimal training data, and a beta
distribution is usually an effective estimate of the PIT distribution: because a 
beta density can be either convex or concave, it is flexible enough to fit the
PIT distribution of overconfident and underconfident forecasters; and because  
the mean can be in the interval $(0,1)$, it can fit biased forecasters
as well. However, problematic behaviors arise at the tails. Except in
exceptional cases (one or both of its two shape parameters is exactly 1), the
beta density is 0 or $\infty$ at the endpoints of its support, which can cause
problmes for recalibration (there can be a big gap between the true PIT density
and \smash{$\hat{g}$} in the tails).

\subsection{Null correction}
The final component of the recalibration ensemble is a null correction, in which
there is no recalibration at all, i.e., we simply set $f_i^*(y) = f_i(y)$. This
prevents overfitting and decreases variance of the overall ensemble correction,
to be described next.  

\subsection{Recalibration ensemble}
The final recalibration system uses the three components described previously
and weights them in an ensemble. The ensemble weights are calculated to
maximize the overall log score. Letting $f^*_{ij}$ denote the forecast density
for sample $i$ and component $j$, the weights ensemble $w$ are defined by
solving the optimization problem:  
\begin{equation}
\label{eq:ensemble}
\minimize_w \;\; \frac{1}{N} \sum_{i=1}^N \log \Bigg(
\sum_{j=1}^p w_j f^*_{ij}(y_i) \Bigg) \;\; \st \;\; w \geq 0, \; 
\sum_{j=1}^p w_j = 1,
\end{equation}
where $p$ is the number of ensemble components (for us, $p=3$) and 
the constraint $w \geq 0$ is to be interpreted componentwise. 

\subsection{Recalibration under seasonality}
Epidemic forecasting presents a new challenge for recalibration. The methodology 
discussed above assumes that the previous behavior of a forecaster is indicative
of future behavior, or more concretely, that the PIT distribution on the training
set will be similar to that on the test set. However, this is not necessarily
the case in epidemic forecasting, due to the fact that a forecaster's
behavior generally changes across the different phases of an epidemic. For
example, some forecasters are too conservative and do not predict enough of a
change in disease incidence from one week to the next. For such a forecaster,
the PIT values are usually too high between a season's onset and peak, because
incidence increases more quickly than forecasted. Conversely, after the season
peaks, the PIT values are too low, because incidence decreases more quickly than
forecasted. 

In order to account for such nonstationarity in the PIT distribution, and at the
same time take advantage of the seasonal nature of epidemic forecasting, we can  
recalibrate a forecast made in a particular week $i$ of a given epidemic season
by forming and using a special training set based on forecasts made at nearby 
weeks in different seasons. For example, a forecast made in week 6 can be
recalibrated based on forecasts in other seasons made in weeks in between 3 and
9. This is what we do in our experiments in this paper, with more details given
in the next section. 

\section{Results}
We apply this ensemble recalibration method to data from influenza forecasting
in the US. In an effort to better prepare for seasonal influenza, the US CDC has
organized a seasonal influenza forecasting challenge every year since 2013,
called the FluSight Challenge \cite{flusight}. In 2017, a group of forecasters
formed the FluSight Network \cite{reich2019b} and began submitting an ensemble
forecast of 27 component forecasters. As part of this collaboration, each of
these forecasters produced and stored retrospective forecasts spanning 9
seasons, from 2010-11 to 2018-19. These forecasters include mechanistic and 
non-mechanistic forecasters, as well as baseline forecasters. They are diverse
in behavior, accuracy, and calibration, and therefore provide an interesting
challenge for our recalibration method, which treats the forecaster as a black
box.

First, we summarize the retrospective forecasts in the FluSight data set. Each
week, a forecast is produced for seven forecasting targets, all of which are
based on weighted ILI (wILI), a population-weighted average of the percentage of
outpatient visits with influenza-like illness derived from reports to the CDC
from a network of healthcare providers called ILINet \cite{ilinet}. The
forecasting targets are: 
\begin{itemize}
    \item season onset (the first week where wILI is above a predefined
    baseline for three consecutive weeks);
    \item season peak week (week of maximum wILI);
    \item season peak percentage (maximum wILI value);
    \item the wILI value at 1, 2, 3, and 4 weeks ahead of the current week.  
\end{itemize}
The first three targets are referred to as seasonal targets and the last four
targets are referred to as short-term targets. Each forecast is submitted as a
binned probability distribution, in intervals of weeks for season onset and
season peak week, and intervals of 0.1\% wILI for season peak percentage and all
short-term targets. Forecasts are produced for each of the 10 HHS Regions as
well as the US as a whole, for a total of 9 seasons, from 2010-11 to 2018-19.
Thus to be clear, the forecasts in this FluSight data set are indexed by
forecaster, target, season, forecast week, and location.

Next, we describe the training setup we use for recalibrating the forecasts in
this data set, which is a kind of nested leave-one-season-out
cross-validation. This is laid out in the steps below for a given forecaster and
forecasting target, and a particular season $s$. 

\begin{enumerate}
    \item Create recalibrated forecasts for all seasons $r \neq s$, using each
      of the three methods: nonparametric, parametric, and null. For a
      forecast in season $r$ at week $i$ and at location $\ell$, we build a
      training set using PIT values from all seasons other than $r$ and $s$, all
      forecast weeks in $[i-3, i+3]$ (within three weeks of $i$), and all
      locations.   
    \item Optimize the ensemble weights $w$ by solving \eqref{eq:ensemble}  
      using the recalibrated forecasts from Step 1.
    \item Create recalibrated forecasts for season $i$, again using each of the
      three methods: nonparametric, parametric, and null. This is just as in
      Step 1, except we use one more season in the training set. Explicitly, for
      a forecast in season $s$ at week $i$ and at location $\ell$, we build a 
      training set using PIT values from all seasons other than $s$, all
      forecast weeks in $[i-3, i+3]$ (within three  weeks of $i$), and all
      locations.    
    \item Create ensemble recalibrated forecasts from season $i$, using the
      recalibration components from Step 3 and the weights from Step 2.
\end{enumerate}

In what follows, we present and discuss the results. The code and data used to
produce all of these results is publicly available online \cite{rumack2021}. 

\subsection{Effect of varying window size}
\begin{figure}[tb]
    \centering
    \includegraphics[width=0.85\textwidth]{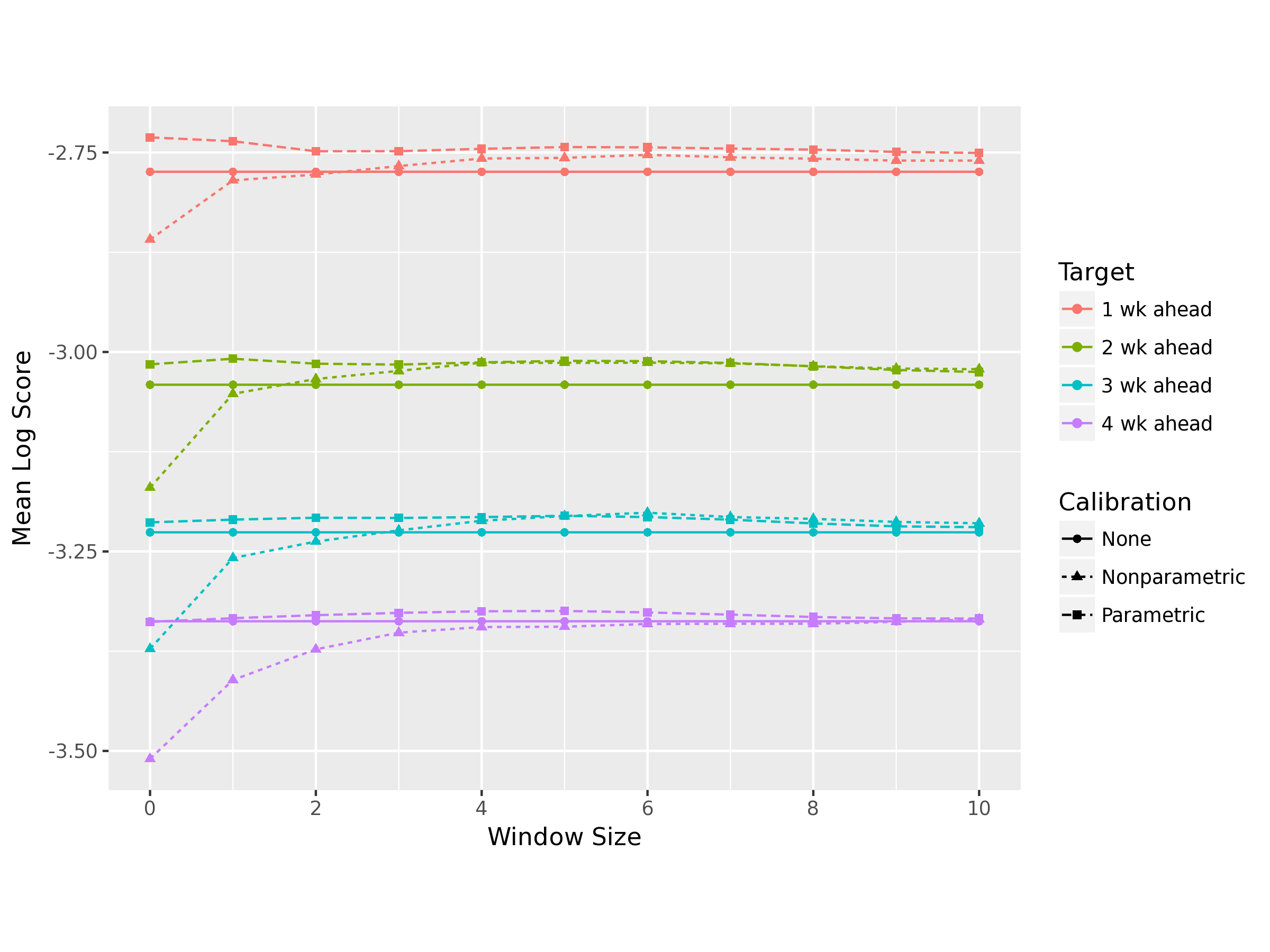}
    \caption{\small Mean log score for the most accurate forecaster of the
      short-term targets, for the different recalibration methods. A window
      size of $k$ corresponds to training recalibration on forecasts within $k$
      weeks of the given forecast week, inclusive. Log score is averaged over 
      9 seasons, 11 locations, and 29 weeks (higher log score is better). The
      window size hardly affects performance of the parametric recalibration
      model. 
      However, the smallest window sizes hurt the nonparametric model.}
    \label{fig:window}
\end{figure}

The training procedure just presented assumes a window of $k=3$ weeks on 
either side of a given week $i$ in order to build the set of PIT values used for
recalibration (using forecast data from other seasons). However, we could
consider varying $k$, which would navigate something like a bias-variance
tradeoff. We would expect the optimal window $k$ to be larger for the
nonparametric recalibration method versus the parametric one. It turns out that 
$k=3$ is typically a reasonable choice for both, as displayed in
Fig~\ref{fig:window}, an example using data from the forecaster scoring highest
on the short-term targets.  

\subsection{Forecast accuracy and calibration}
\begin{figure}[p]
    \centering
    \includegraphics[width=0.85\textwidth]{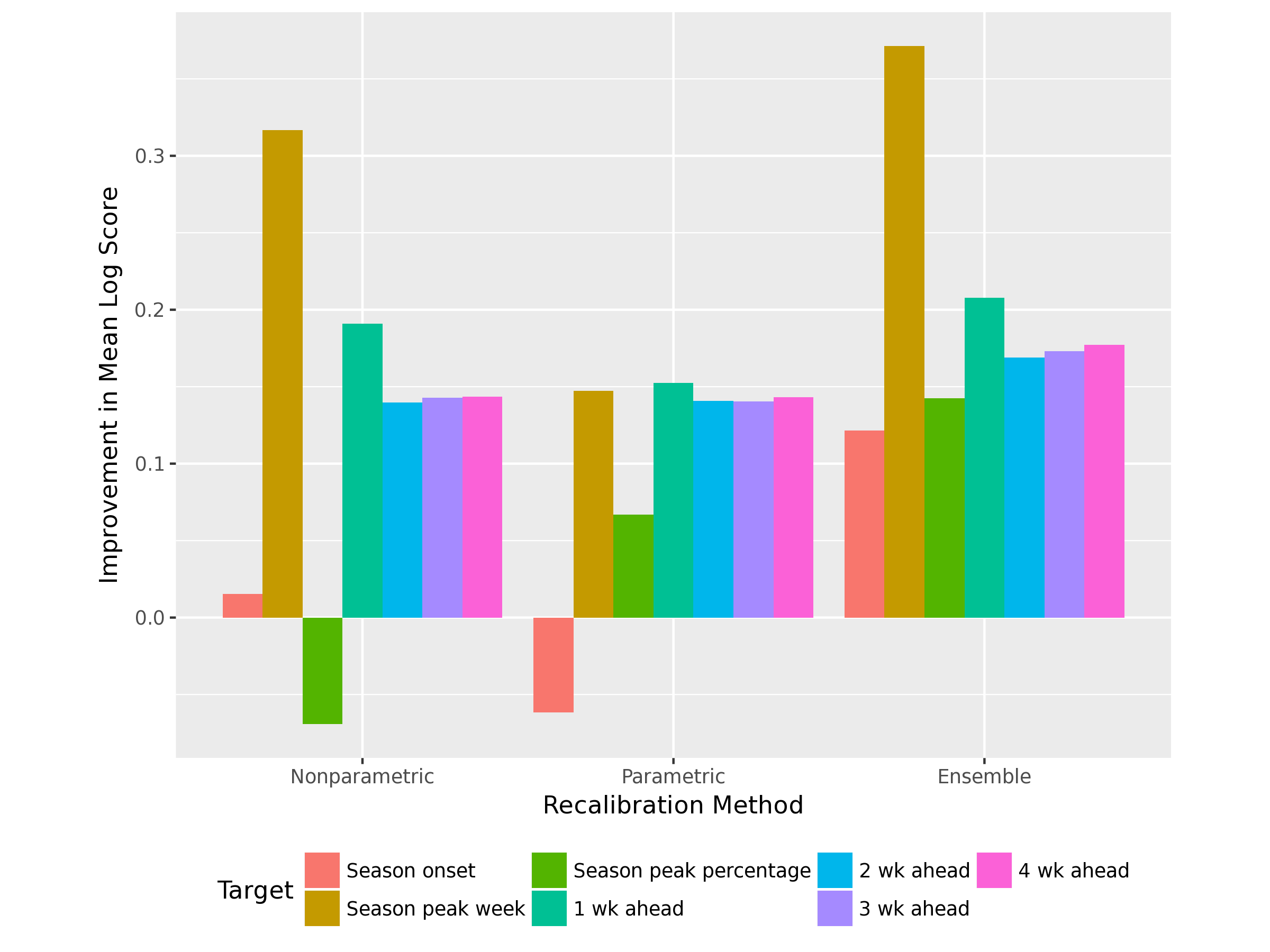}
    \caption{\small Improvement in mean log score, for the different
      recalibration methods. Log score is averaged over all 27 forecasters in
      the FluSight, 9 seasons, 11 locations, and 29 weeks (higher log score is
      better). The ensemble recalibration method improves accuracy for every
      target.}
    \label{fig:mls_improvement}

    \bigskip
    \includegraphics[width=0.85\textwidth]{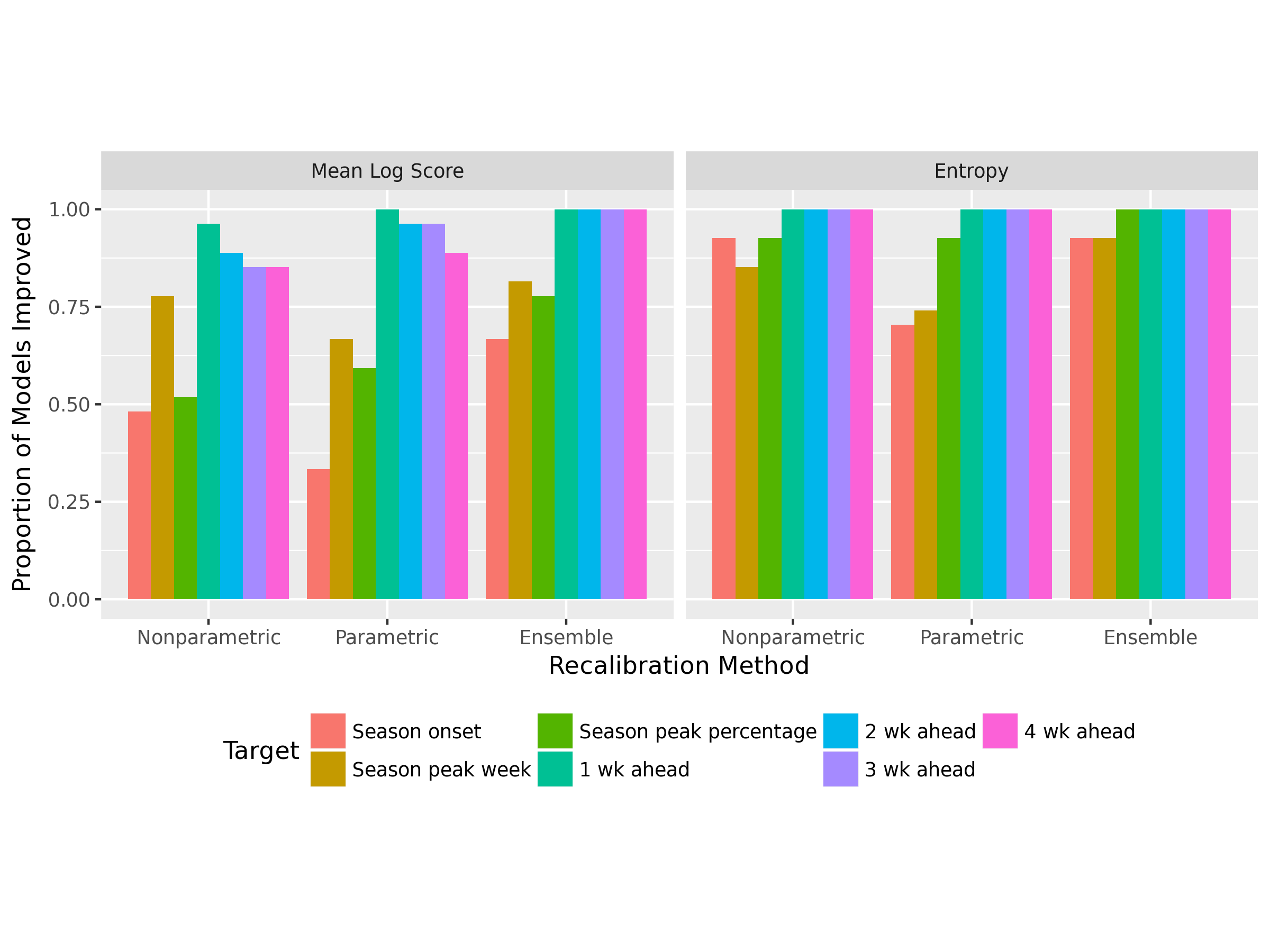}
    \caption{\small Proportion of forecasters for which recalibration improves
      mean log score (left) and entropy of the PIT values (right). The ensemble
      method improves accuracy for the short-term targets for all forecasters,
      and most forecasters for the seasonal targets. It also improves
      calibration (as measured by entropy) for most forecasters and most
      targets. The ensemble method outperforms both the nonparametric and
      parametric methods.}  
    \label{fig:model_improvement_mls_entropy}
\end{figure}

For the short-term targets, the ensemble recalibration method improves
the mean log score for almost all forecasters. Both the nonparametric
and parametric recalibration methods significantly improve the mean log
score, and the ensemble improves it even further. For the seasonal
targets, some component recalibration methods do not improve accuracy,
although the ensemble method does improve accuracy, averaged over all
forecasters. However, the ensemble improves accuracy for seasonal
targets in only about three-quarters of forecasters. See
Fig~\ref{fig:mls_improvement} and Fig~\ref{fig:model_improvement_mls_entropy}.

\begin{figure}[p]
    \centering
    \includegraphics[width=0.8\textwidth]{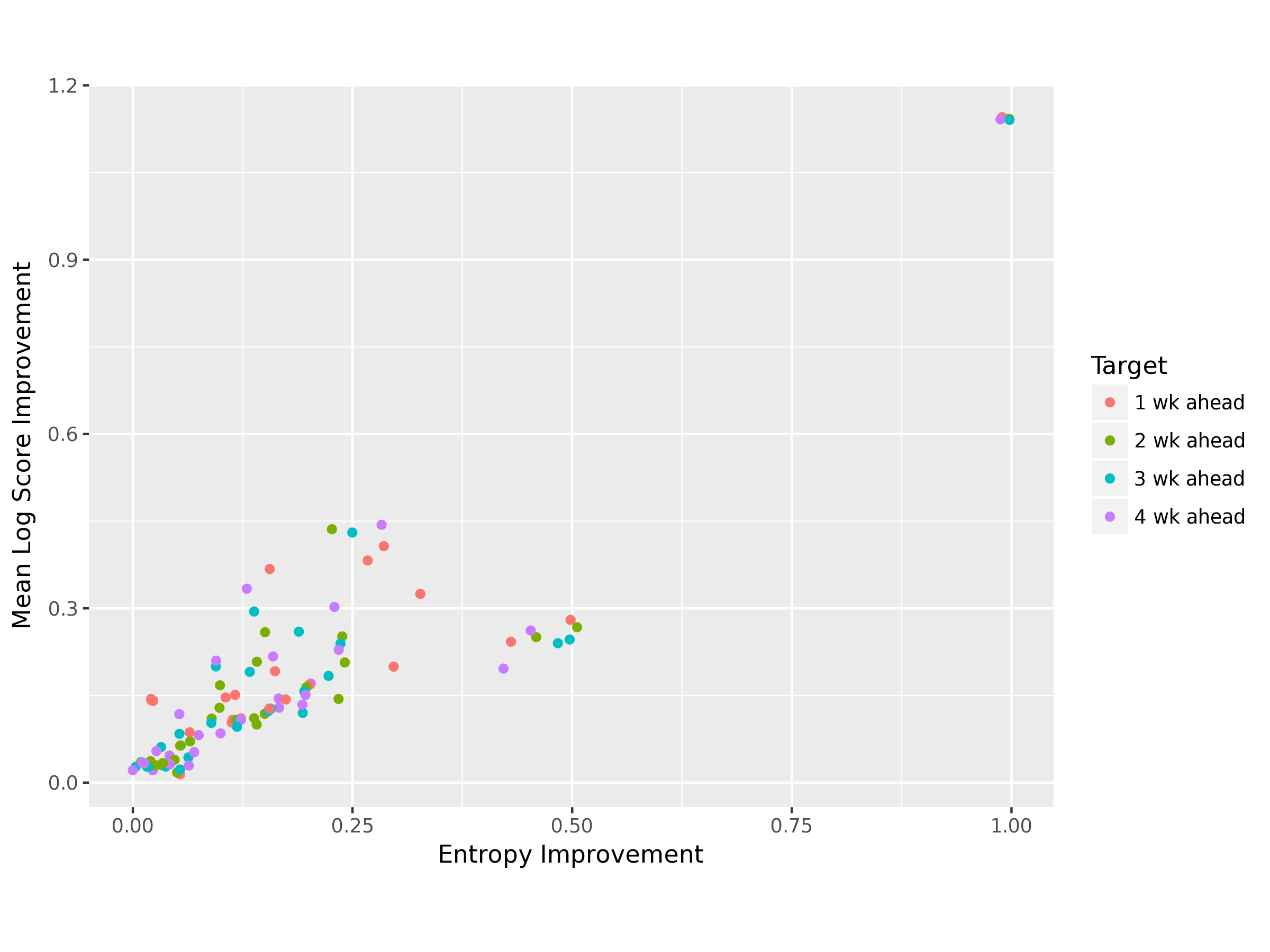}
    \caption{\small Improvement in mean log score versus improvement in entropy  
      for each of the 27 FluSight forecasters and short-term targets. There is a
      clear linear trend (with slope approximately 1) between the improvement in
      calibration and the improvement in accuracy.}  
    \label{fig:short_term_ensemble_improvement}

    \bigskip
    \includegraphics[width=0.8\textwidth]{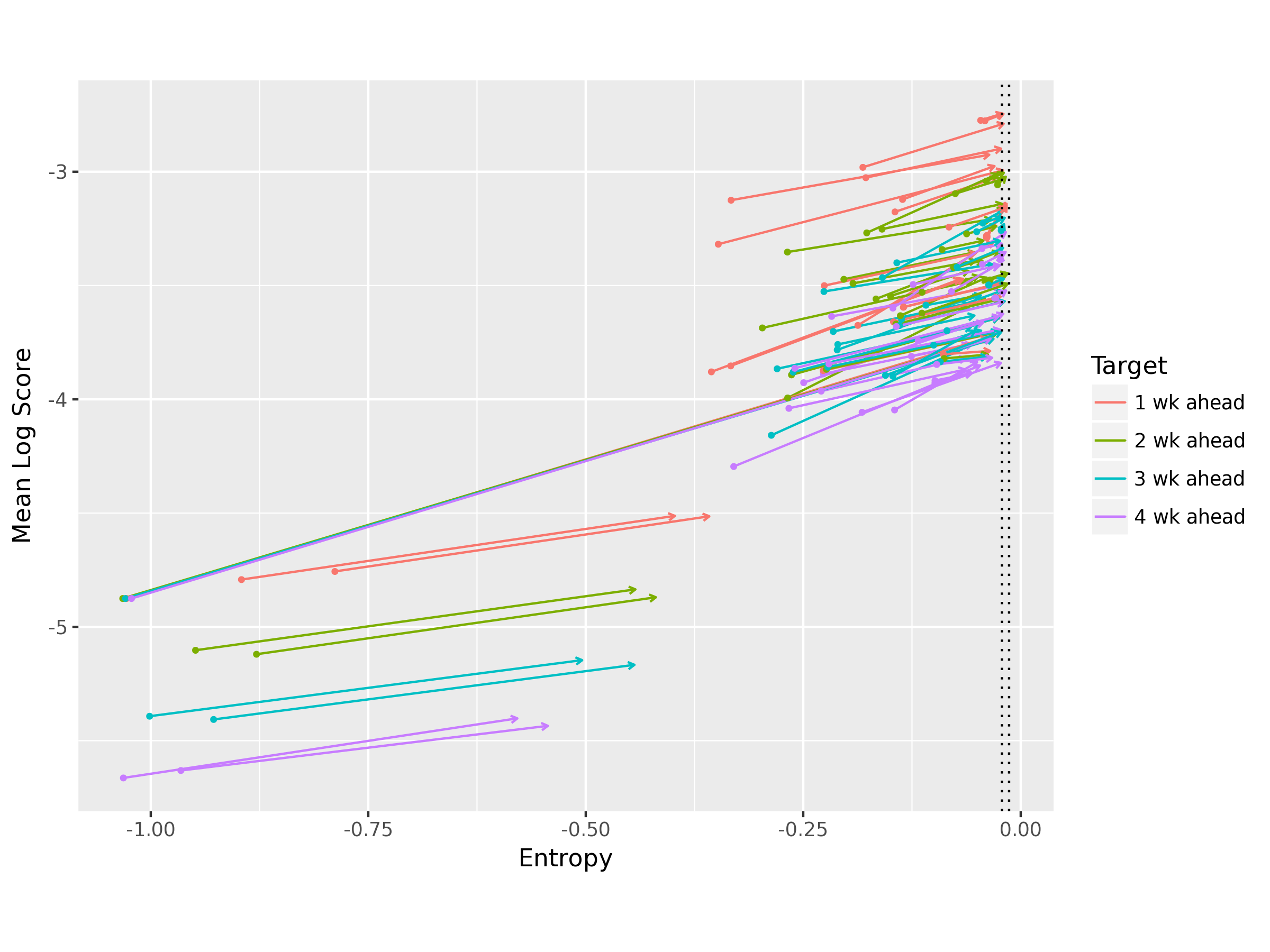}
    \caption{\small Entropy and mean log score before and after recalibration,
      for each of the 27 FluSight forecasters and short-term targets. The tail
      of arrow represents a quantity before recalibration, and the head after
      recalibration. The dotted lines show the central 90\% interval of the
      entropy of a comparably-sized sample of standard uniform random variables
      for  comparison. For all but two forecasters (the eight line bottom-most
      line segment), the ensemble recalibration method achieves  almost perfect
      calibration as evidenced by a near-zero PIT entropy, and this is
      accompanied by significant improvements in accuracy.} 
    \label{fig:improvement_st}
\end{figure}

Furthermore, in Fig~\ref{fig:model_improvement_mls_entropy} we see that 
the recalibrated forecaster's improvement in calibration (as measured by
entropy) is quite close to the improvement in mean log score. This confirms our
expectations from \eqref{eq:calibration_and_log_score}. 
Fig~\ref{fig:short_term_ensemble_improvement} gives a more direct comparison of
improvements in accuracy versus calibration, i.e., in mean log score versus
entropy, for the short-term forecasts. (Note that we estimate the entropy of the  
distribution of PIT values using a simple histogram estimator with 100 equal
bins along the interval $[0,1]$.) We see a clear linear trend, with slope
approximately 1, again confirming \eqref{eq:calibration_and_log_score}.   

Finally, in Fig~\ref{fig:improvement_st}, we show that our ensemble
recalibration method increases the entropy of the PIT distribution to nearly
zero for nearly every forecaster. The two exceptions, the line segments towards 
the bottom of Fig~\ref{fig:improvement_st}, correspond to particularly poor
forecasters (so poor that are outperformed by a baseline forecaster that outputs
a uniform distribution).   

\subsection{Recalibrating the FluSight ensemble} 
As we just saw, recalibration improves the performance of the individual
forecasters in the FluSight Network. A natural follow up is therefore to
investigate whether it can improve the performance of the FluSight ensemble,
a forecaster that combines 27 component forecasters (the individual FluSight
forecasters), whose construction is described in \cite{reich2019b}. 

As both recalibration and ensembling are post-processing methods (i.e., that can
be applied in post-processing of forecast data), we are left with two options to
explore. We can recalibrate the component forecasters and then ensemble (C-E),
or ensemble the components and then recalibrate (E-C). In the C-E model, we
train ensemble weights in a leave-one-season-out format, on the recalibrated
component forecasts. In the E-C model, we train ensemble weights in a
leave-one-season-out format on the original component forecasts, and then
recalibrate the ensemble forecasts. 

\begin{figure}
    \centering
    \includegraphics[width=0.85\textwidth]{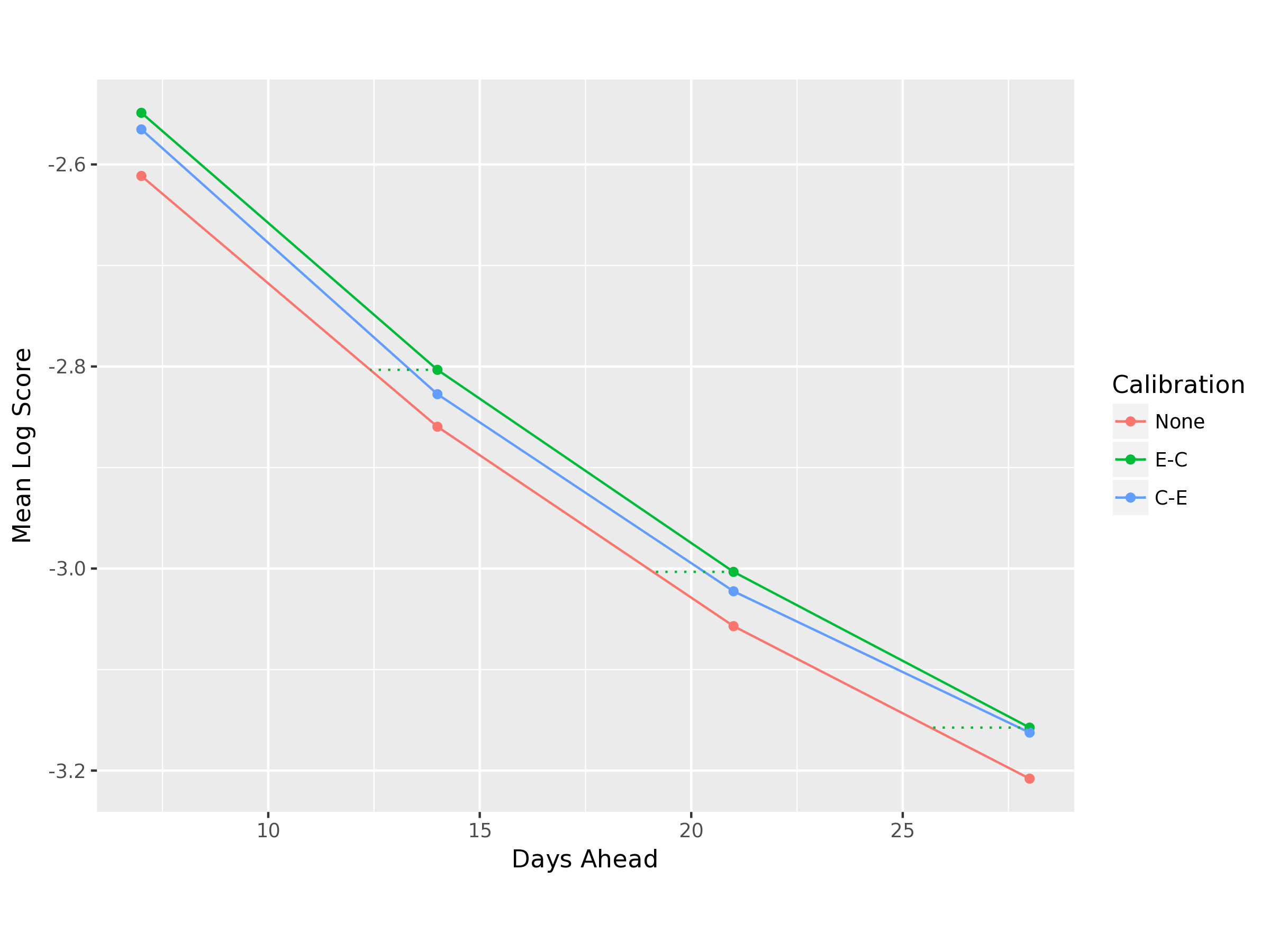}
    \caption{\small Mean log score for the two different approaches to
      recalibrating the FluSight ensemble forecaster, with C-E and E-C
      reflecting the order of recalibration and ensembling. Both the C-E and E-C
      models outperform the original ensemble (with no recalibration), but
      ensembling followed by recalibration performs best. By viewing forecast
      performance as a function of time, recalibration increases performance as
      much as roughly two days' time would.} 
    \label{fig:ensemble_cal}
\end{figure}

Fig~\ref{fig:ensemble_cal} reveals that E-C model performs better than the C-E
model. This is in line with established forecasting theory, which states that
linear ensembles (which take a linear combination of component forecasters, such
as the FluSight ensemble approach) themselves are generally miscalibrated, even
when the individual component forecasters are themselves calibrated
\cite{hora2004, gneiting2010, gneiting2013}.

\section{Discussion}
Even in a domain as complex as epidemic forecasting, relatively simple
recalibration methods such as those described in this paper can significantly
improve both calibration and accuracy. A forecaster's performance for any proper
score can be decomposed into three components: the inherent uncertainty of the
target itself, the resolution of the forecaster (concentration of the
forecasts), and the reliability of the forecaster to the target (calibration)
\cite{brocker2009}. In epidemic forecasting, without seasonality-aware
recalibration training (such as that proposed and implemented in this paper),
recalibration will not affect the resolution term, which is left to the
individual forecasters, but it will improve the reliability term. However, using
seasonality-aware recalibration, it can also improve the resolution term.

Over 9 seasons of forecast data from 27 forecasters in the FluSight Challenge,
we found that recalibration was especially helpful for the short-term targets
(1-4 week ahead forecasts). With the exception of two very similar forecasters
that have poor performance, the ensemble recalibration method was able to reduce
the entropy of the PIT distribution to nearly zero (not or barely statistically
significantly different than a uniform distribution). The recalibrated forecasts
are therefore more accurate and more reliable. This is true across a diverse set
of forecasters, including mechanistic, statistical, baseline, and ensemble
models; indeed, as our recalibration method treats the forecaster as a black
box, it can be applied to any forecaster, given access to suitable training data 
(retrospective historical forecasts).

The performance of recalibration with respect to the seasonal targets (onset,
peak week, and peak percentage) was less conclusive. Although the mean log score 
averaged over all of the forecasters was improved, recalibration only improved 
the performance of about three-quarters of the forecasters. Seasonal targets are  
inherently more difficult to recalibrate because at the end of the season, the
true value has almost certainly been observed, and the forecasts are highly 
confident. For these forecasts, the correct bin has a mass of almost 1, and the
observed PIT value then is approximately 0.5. At the end of the season, the PIT
distribution is very concentrated at 0.5, which indicates underconfidence and
poor calibration. If these PIT values of 0.5 are used to train forecasts for
recalibration earlier in the season, before the target is observed, then
recalibration incorrectly makes the forecast more confident. Because one is
unsure whether the season peak has occurred or not for several weeks after the
peak occurs, recalibration training is a nontrivial task. In general, more work
is required to reliably improve accuracy and calibration for seasonal targets,
which is a topic for future work. 

\section*{Acknowledgements}
AR was supported by a fellowship from the Center for Machine Learning at
Carnegie Mellon University and a gift from the McCune Foundation. RT and RR were
support by Centers for Disease Control and Prevention grant U01IP001121.

\nolinenumbers
\bibliography{bib.bib}

\end{document}